\documentclass[10pt,twocolumn,letterpaper]{article}

\usepackage[pagenumbers]{cvpr} %
\usepackage{comment}

\definecolor{cvprblue}{rgb}{0.21,0.49,0.74}
\usepackage[pagebackref,breaklinks,colorlinks,allcolors=cvprblue]{hyperref}

\usepackage{amsmath}
\usepackage{amssymb}
\usepackage{mathtools}
\usepackage{amsthm}
\usepackage{multirow}
\usepackage{multicol}
\usepackage{makecell}
\usepackage{colortbl}
\usepackage{xcolor}
\usepackage{wrapfig}

\renewcommand{\paragraph}[1]{\vspace{1mm}\noindent\textbf{#1}}
\newlength\savewidth\newcommand\shline{\noalign{\global\savewidth\arrayrulewidth
  \global\arrayrulewidth 1pt}\hline\noalign{\global\arrayrulewidth\savewidth}}
\newcolumntype{x}[1]{>{\centering\arraybackslash}p{#1pt}}
\newcolumntype{y}[1]{>{\raggedright\arraybackslash}p{#1pt}}
\newcolumntype{z}[1]{>{\raggedleft\arraybackslash}p{#1pt}}
\definecolor{mygreen}{RGB}{0, 205, 108}
\definecolor{convcolor}{HTML}{412F8A}
\definecolor{resnetcolor}{HTML}{8DA0CB}
\definecolor{vitcolor}{HTML}{fc8e62}

\hypersetup{
    colorlinks=true,
    linkcolor=red,
    citecolor=blue,
    filecolor=magenta,      
    urlcolor=magenta,
    }
\newcommand{\tablestyle}[2]{\setlength{\tabcolsep}{#1}\renewcommand{\arraystretch}{#2}\centering\footnotesize}

\definecolor{orange}{HTML}{ff7f0e}
\definecolor{blue}{HTML}{1f77b4}
\definecolor{baselinecolor}{gray}{.9}
\newcommand{\baseline}[1]{\cellcolor{baselinecolor}{#1}}

\def\paperID{11645} %
\def\confName{CVPR}
\def\confYear{2025}

\title{Neural Network Diffusion}

\author{
Kai Wang$^{1}$\hspace{.5em} Dongwen Tang$^{1}$\hspace{.5em} Boya Zeng$^{2}$\hspace{.5em} Yida Yin$^{3}$\hspace{.5em}
Zhaopan Xu$^{1}$ \\ Yukun Zhou$^{1}$\hspace{.5em} Zelin Zang$^{1}$\hspace{.5em} Trevor Darrell$^{3}$\hspace{.5em} Zhuang Liu$^{4*}$\hspace{.5em} Yang You$^{1*}$ \\[4pt] 
$^{1}$National University of Singapore\hspace{.5em} $^{2}$University of Pennsylvania \\
$^{3}$University of California, Berkeley\hspace{.5em} $^{4}$Meta FAIR
}

\begin{document}
\maketitle
\begin{abstract}
Diffusion models have achieved remarkable success in image and video generation. In this work, we demonstrate that diffusion models can also \textit{generate high-performing neural network parameters}.
Our approach is simple, utilizing an autoencoder and a diffusion model. The autoencoder extracts latent representations of a subset of the trained neural network parameters. Next, a diffusion model is trained to synthesize these latent representations from random noise. This model then generates new representations, which are passed through the autoencoder's decoder to produce new subsets of high-performing network parameters. Across various architectures and datasets, our approach consistently generates models with comparable or improved performance over trained networks, with minimal additional cost. Notably, we empirically find that the generated models are not memorizing the trained ones. Our results encourage more exploration into the versatile use of diffusion models.
Our code is available \href{https://github.com/NUS-HPC-AI-Lab/Neural-Network-Diffusion}{here}.
\end{abstract}

\newcommand\blfootnote[1]{%
  \begingroup
  \renewcommand\thefootnote{}\footnote{#1}%
  \addtocounter{footnote}{-1}%
  \endgroup
}
\blfootnote{\hspace{-1em} $^*$equal advising}

\section{Introduction}

The origin of diffusion models can be traced back to non-equilibrium thermodynamics~\citep{jarzynski1997equilibrium, sohl2015deep}. Diffusion was first applied to progressively remove noise from inputs and generate clear images in \cite{sohl2015deep}. Later works, such as DDPM~\citep{ho2020denoising} and DDIM~\citep{song2021denoising}, refined diffusion models with a training paradigm characterized by forward and reverse processes.

At that time, diffusion models were not yet capable of generating high-quality images.
Guided-Diffusion~\citep{dhariwal2021diffusion} enhanced diffusion models with a more effective architecture and thereby surpassed GAN-based methods~\citep{zhu2017unpaired, isola2017image}. Since then, GLIDE~\citep{nichol2021glide}, Imagen~\citep{saharia2022photorealistic}, DALL$\cdot$E 2~\citep{ramesh2022hierarchical}, and Stable Diffusion~\citep{rombach2022high} have achieved photorealistic image generations, becoming widely adopted by artists and creatives.

Despite the great success of diffusion models in visual generation, their potential in other domains remains largely unexplored. In this work, we demonstrate the surprising capability of diffusion models in \textit{generating high-performing neural network parameters}.
Parameter generation aims to produce neural network parameters that perform well on specific tasks and has been approached through prior and probability modeling, such as stochastic neural network~\cite{sompolinsky1988chaos, wong1991stochastic, schmidt1992feed,murata1994network} and Bayesian neural network~\cite{neal2012bayesian, kingma2013auto, rezende2014stochastic, kingma2015variational}. However, leveraging diffusion models for parameter generation has not been well-explored yet.

\begin{figure}[tp]
\centering
\includegraphics[width=0.95\linewidth]{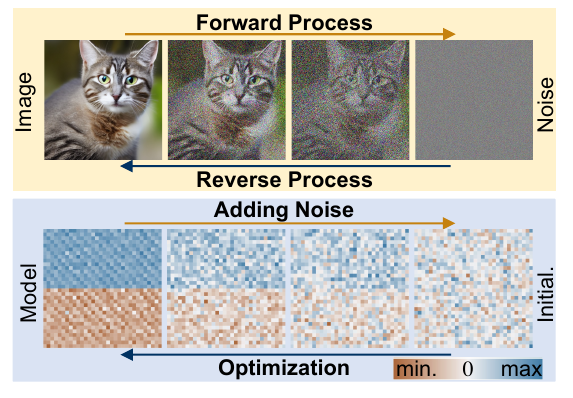}
\caption{
\textit{Top} illustrates the standard diffusion process in image generation.
\textit{Bottom} shows the parameter heatmap of the batch normalization (BN) layer at various stages of ResNet-18 training on CIFAR-100.
In the heatmap, \textit{the upper half} is BN weights, while \textit{the lower half} is BN biases. Color corresponds to parameter value.
} 
\label{fig:motivation}
\end{figure}

Let's take a closer look at neural network training and diffusion-based image generation in Figure~\ref{fig:motivation}. They share two commonalities:
i) Both neural network training and the reverse process of diffusion models transition from random noise or initialization to specific distributions.
ii) High-quality images and well-optimized parameters can likewise be degraded into simple distributions, such as Gaussian distribution, through multiple rounds of noise addition.

Based on the observations above, we introduce a novel approach for parameter generation, named Neural Network Diffusion (\textbf{p-diff}, where `p' stands for parameter), which employs a standard latent diffusion model to synthesize new sets of parameters.
Our method is simple, comprising an autoencoder and a standard latent diffusion model to learn the distribution of high-performing parameters. First, we train the autoencoder on a subset of neural network parameters to extract their latent representations. The diffusion model is then optimized to generate these latent representations from random noise. This allows us to synthesize new latent representations from random noise and feed these representations through the trained autoencoder's decoder to produce new, high-performing model parameters.

Our approach has the following characteristics: i) It consistently achieves similar or enhanced performance compared to the training data (\textit{i.e.,} models trained by gradient-based optimizers) across various datasets and architectures.
ii) Our generated models show great differences in predictions compared to the models in training data. Therefore, our approach synthesizes novel, high-performing model parameters rather than memorizing the training samples. We hope our work can motivate further research to expand the applications of diffusion models to other domains.

\section{Neural Network Diffusion}
In this section, we review the preliminaries of diffusion models and introduce the components of our approach.
\subsection{Preliminaries of Diffusion Models}
Diffusion models typically consist of forward and reverse processes in a multi-step chain indexed by timesteps. We introduce these two processes in the following:

\paragraph{Forward process.} Given a sample $x_{0} \sim q(x)$, the forward process progressively adds Gaussian noise for $T$ steps and obtain $x_{1}, x_{2}, \cdots, x_{T}$. The process is formulated as:
\begin{equation}
\begin{split}
    q(x_{t}|x_{t-1}) &= \mathcal{N}(x_{t}; \sqrt{1-\beta_{t}}x_{t-1}, \beta_{t}\mathbf{I}),
\end{split}
\end{equation}
where $\mathcal{N}$ is a Gaussian distribution, and $\beta_{t}$ controls the noise variance at each step $t$, and $\mathbf{I}$ is the identity matrix. Note, for any arbitrary step $t$, we can directly sample:
\begin{equation}
    x_t=\sqrt{\bar{\alpha}_{t}}x_0 + \sqrt{1-\bar{\alpha}_{t}}\epsilon_t,
\end{equation} where $\bar{\alpha}_t=\prod_{s=1}^t(1-\beta_s)$ and $\epsilon_t \sim \mathcal{N}(0, \mathbf{I})$.

\paragraph{Reverse process.} In contrast to the forward process, the reverse process aims to remove the noise from the input $x_{t}$. A neural network, parameterized by $\theta$, is trained to learn this reverse process:
\begin{equation}
\begin{split}
    p_{\theta}(x_{t-1}|x_{t}) &= \mathcal{N}(
    \mu_{\theta}(x_{t}, t), \Sigma \substack{\\ \\ \theta}(x_{t}, t)),
\end{split}
\end{equation}
where $\mu_{\theta}(x_{t}, t)$ and $\Sigma \substack{\\ \\ \theta}(x_{t}, t)$ are the estimated Gaussian mean and variance outputted by the denoising network. 

\paragraph{Training and inference.} The model $p_\theta$
is trained  with the variational lower bound~\cite{kingma2013auto} of the log-likelihood of $x_0$. This training objective is expressed as:
\begin{equation}
    \mathcal{L}(\theta) = \sum_t \mathcal{D}_{\text{KL}}(q(x_{t-1}|x_{t}, x_{0})||p_{\theta}(x_{t-1}|x_{t})),
\label{eq:loss}
\end{equation}
where the $ \mathcal{D}_{\text{KL}}(\cdot||\cdot)$ denotes the Kullback–Leibler (KL) divergence between two distributions.

During the inference, we first sample a random Gaussian noise $x_t \sim \mathcal{N}(0,\mathbf{I})$ and then obtain $x_{t-1} \sim p_\theta(x_{t-1} | x_t)$ recursively until we get a clean output.

\begin{figure*}[h]
\vspace{-.5em}
\centering
\includegraphics[width=0.75\linewidth]{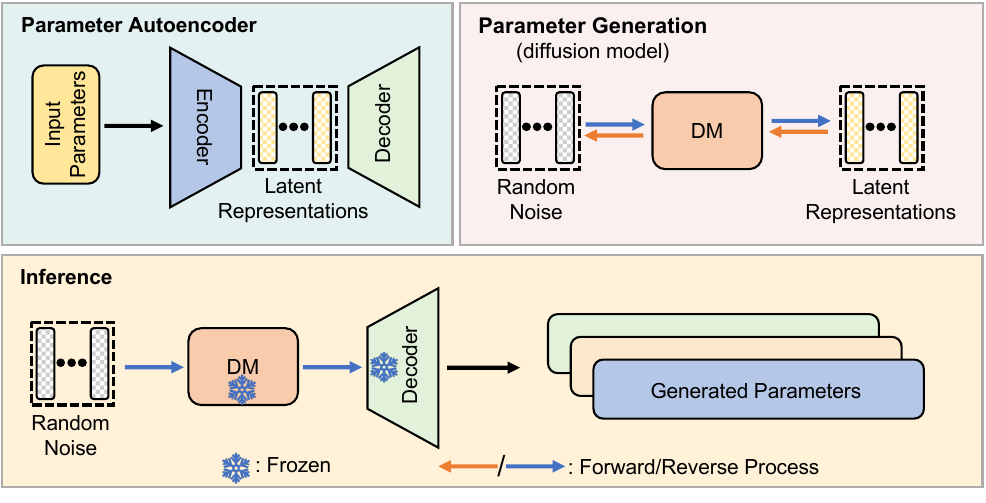}
\vspace{-.5em}
\caption{
Our approach consists of two processes: parameter autoencoder and parameter generation. Parameter autoencoder aims to extract the latent representations and reconstruct model parameters via the decoder.
The extracted representations are used to train a diffusion model (DM).
During inference, a random noise vector is fed into the DM and the trained decoder to generate new parameters.}
\label{fig:pipeline}
\end{figure*}

\subsection{Overview}
We propose neural network diffusion (p-diff) to generate high-performing neural network parameters from random noise.
As illustrated in Figure~\ref{fig:pipeline}, our method consists of two processes: parameter autoencoder and parameter generation.
Given a set of high-performing model checkpoints, we first select a particular subset of parameters from every model and flatten them into 1-dimensional vectors.
Subsequently, we use an encoder to extract latent representations from these vectors, and a decoder to reconstruct the parameters from the latent representations.
Then, a diffusion model is trained to synthesize latent representations from random noise.
After training, we use p-diff to generate new parameters via the following chain: random noise $\rightarrow$ reverse process $\rightarrow$  trained decoder $\rightarrow$ generated parameters.

\subsection{Parameter Autoencoder}
\label{sec:param_autoencoder}
\paragraph{Data preparation.}
To collect the training data for the autoencoder, we train a model on the target task until convergence and save a checkpoint at each of 300 additional training steps.
In our paper, we default to synthesizing a subset of model parameters. Thus, in the 300 additional training steps, we only update the selected subset of parameters via a gradient-based optimizer and fix the remaining parameters of the model.
The saved subsets of parameters $S = [s_{1},\ldots, s_{k},\ldots, s_{K}]$ are then used to train the autoencoder, where $K$ is the number of the training samples.

\paragraph{Training.} We flatten the subset of parameters $s_{k}$ into 1-dimensional vector $v_{k} \in \mathbb{R}^{d}$
, where $d$ is the size of the selected subset of parameters. After that, an autoencoder is trained to reconstruct these parameters. 
To enhance the robustness and generalization of the autoencoder, we add Gaussian noise to both the parameter vector $v_k$ and the latent representation $z_k$ as augmentation, and control the noise scale with $\sigma^2_v$ and $\sigma^2_z$. Denote the noise on parameter vector as $\xi_{v_k}$ and the noise on latent representation as $\xi_{z_k}$. The entire encoding and decoding proceses are:
\vspace{-2pt}
\begin{equation}
    \begin{aligned}
        z_{k} = \underbrace{f_{\phi}(v_{k}+\xi_{v_{k}})}_{\text{encoding}}, \xi_{v_{k}}\sim\mathcal{N}(0, \sigma^2_v\mathbf{I});  \\
    \widehat{v_{k}} = \underbrace{f_{\pi}(z_{k}+\xi_{z_{k}})}_{\text{decoding}}, \xi_{z_{k}}\sim\mathcal{N}(0, \sigma^2_z\mathbf{I}),
    \end{aligned}
\end{equation}
\vspace{-2pt}where $f_{\phi}(\cdot)$ and $f_{\pi}(\cdot)$ denote the encoder and decoder parameterized by $\phi$ and $\pi$, respectively, and $\widehat{v_{k}}$ is the reconstructed output from the decoder.
To train our parameter autoencoder, we use the standard objective of minimizing the $L^2$ norm between $\widehat{v_{k}}$ and $v_{k}$.

\subsection{Parameter Generation}
\label{sec:pg}
\textbf{Training and inference.} One direct strategy to approach parameter generation is to train a diffusion model directly on parameter space.
However, the memory cost of this is too heavy, especially when the dimension of the parameters to generate is very large. Therefore, we apply the diffusion process to the latent representations instead. Based on the latent representation $z_{k}$ extracted from our parameter autoencoder, we can apply the reparamterization trick proposed in DDPM~\citep{ho2020denoising} to rewrite Equation~\ref{eq:loss}. Then the training objective for the denoising network simplifies to a mean-squared error:
\begin{equation}
    \mathcal{L}(\theta) = ||\epsilon -\epsilon_{\theta}(\sqrt{\bar{\alpha}_{t}}z_{k} + \sqrt{1-\bar{\alpha}_{t}}\epsilon, t)||^{2},
\end{equation}
where $\epsilon_{\theta}(\sqrt{\bar{\alpha}_{t}}z_{k} + \sqrt{1-\bar{\alpha}_{t}}\epsilon, t)$ is the predicted noise from the network $p_\theta$ and $\epsilon_t$ is the ground truth sampled Gaussian noise. Once the noise predictor $p_\theta$ has been trained, we directly feed random noise into the reverse process and the trained decoder to generate a new set of high-performing parameters. During evaluation, these generated parameters are concatenated with the remaining parameters in the original model to form a new one.

\paragraph{Design space.} Neural network parameters differ from image pixels in several ways, including data type, dimensions, range, and physical interpretation. 
Unlike images, neural network parameters generally lack spatial structure, so we use pure 1D convolutions instead of 2D convolution in our parameter autoencoder and diffusion model.

\section{Experiments}
\vspace{-.5em}
\begin{table*}[tp]
\centering
\tablestyle{2.5pt}{1.2}
\begin{tabular}{l|ccc|ccc|ccc|ccc|ccc|ccc}
& \multicolumn{3}{c|}{CIFAR-100} & \multicolumn{3}{c|}{STL-10} & \multicolumn{3}{c|}{Flowers} & \multicolumn{3}{c|}{Pets} & \multicolumn{3}{c|}{Food-101} & \multicolumn{3}{c}{ImageNet-1K} \\
& orig. & ensem. & p-diff 
& orig. & ensem. & p-diff 
& orig. & ensem. & p-diff 
& orig. & ensem. & p-diff 
& orig. & ensem. & p-diff 
& orig. & ensem. & p-diff \\
\shline
ResNet-18 & 77.4 & 77.7 & \textbf{77.9} & 95.1 & 95.1 & \textbf{95.2} & \textbf{80.1} & 79.2 & 79.7 & 89.0 & 88.8 & \textbf{89.1} & 79.1 & 79.4 & 79.4 & 69.5 & 70.3 & \textbf{70.4} \\

ResNet-50 & 78.4 & 78.6 & \textbf{78.7} & 97.0 & 96.9 & \textbf{97.1} & 85.9 & 85.8 & \textbf{86.1} & \textbf{92.6} & 92.1 & 92.5 & \textbf{80.4} & 80.0 & 80.1 & 78.3 & 78.5 & 78.5 \\

ViT-Tiny & 86.7 & 86.4 & 86.7 & \textbf{97.8} & 97.6 & 97.7 & 87.6 & 87.5 & 87.6 & \textbf{90.5} & 89.9 & 90.4 & 87.4 & 87.4 & \textbf{87.5} & 75.2 & 75.4 & 75.4 \\

ViT-Base & 91.2 & 91.1 & 91.2 & 99.1 & 99.1 & 99.1 & 98.1 & 98.0 & \textbf{98.2} & 94.1 & 94.1 & \textbf{94.4} & \textbf{91.3} & 91.2 & 91.2 & 85.1 & 85.1 & 85.1 \\

ConvNeXt-Tiny & 89.8 & 89.6 & 89.8 & \textbf{99.1} & 98.9 & 99.0 & 98.2 & 98.2 & \textbf{98.3} & \textbf{94.2} & 93.7 & 93.9 & 91.6 & 91.5 & 91.6 & 83.4 & 83.7 & 83.7 \\

ConvNeXt-Base & 93.2 & 93.1 & 93.2 & 99.7 & 99.6 & 99.7 & 99.5 & 99.5 & 99.5 & \textbf{95.2} & 94.7 & 95.0 & 92.8 & 92.9 & \textbf{93.0} & 85.1 & 85.3 & 85.3
\end{tabular}
\vspace{-.5em}
\caption{Partial parameters generated by p-diff match or exceed original models' performance.
\textbf{Bold entries} are best results.}
\label{tab:main}
\end{table*}

\begin{table*}[h]
\centering
\tablestyle{2.5pt}{1.2}
\begin{tabular}{l|ccc|ccc|ccc|ccc|ccc}
 & \multicolumn{3}{c|}{ConvNet-mini} & \multicolumn{3}{c|}{MLP-mini} & \multicolumn{3}{c|}{ResNet-mini} & \multicolumn{3}{c|}{ViT-mini} & \multicolumn{3}{c}{ConvNeXt-mini}\\
& original  & ensemble & p-diff  & original  & ensemble & p-diff  & original  & ensemble & p-diff  & original  & ensemble & p-diff  & original  & ensemble & p-diff \\
\shline
CIFAR-10 & \textbf{56.0} & 55.6 &55.8 & 41.8 & 41.8 & \textbf{42.4} & \textbf{58.6}  & 58.1 & 58.5  & 73.4 & 73.6 & 73.6 & \textbf{72.0} & 71.7 & 71.9\\
STL-10 & \textbf{46.4} & 45.6 & 46.1 & 35.0 & 35.0 & \textbf{35.2} & 51.5 & 51.4 & \textbf{51.8} & 52.3 & 52.0 & \textbf{52.6} & 48.1 & 47.8 & 48.1
\end{tabular}
\vspace{-.5em}
\caption{
The best validation accuracy comparison between original models and full parameters generated by p-diff.
P-diff demonstrates strong generality across five architectures (ConvNet-mini, MLP-mini, ResNet-mini, ViT-mini, and ConvNeXt-mini) with number of parameters ranging from 24K to 81K. Each architecture has a distinct structural design. \textbf{Bold entries} indicate best results.}
\label{tab:all_para}
\end{table*}

In this section, we introduce our experimental setup and then report our results and ablation analysis.

\subsection{Setup}\label{sec:exp-setting}
\paragraph{Datasets and architectures.} We evaluate p-diff across a wide range of datasets, including CIFAR-10/100~\citep{krizhevsky2009learning}, STL-10~\citep{coates2011analysis}, Flowers~\citep{nilsback2008automated}, Pets~\citep{parkhi2012cats}, Food-101~\citep{bossard2014food}, and ImageNet-1K~\citep{deng2009imagenet}. To evaluate p-diff's ability to generate new subsets of high-performing network parameters, we conduct experiments on ResNet-18/50~\citep{he2016deep}, ViT-Tiny/Base~\citep{dosovitskiy2020image}, ConvNeXt-Tiny/Base~\citep{liu2022convnet}. Furthermore, to demonstrate our method's capability to generate full neural network parameters, we also experiment on 5 small hand-designed models: ConvNet-mini, MLP-mini, ResNet-mini, ViT-mini, and ConvNeXt-mini. The architectures of these models are detailed in the Appendix.

\paragraph{Training details.}
The parameter autoencoder and latent diffusion model both use encoder-decoder architectures with five 1-dimensional CNN layers and one fully-connected layer.
For the ImageNet-1K dataset, we take the \emph{pretrained} model from the timm library~\cite{rw2019timm}; for other smaller datasets, we train all model parameters on each dataset until accuracy converges. Then, we fine-tune the last two normalization layers of the model for 300 training steps, saving one checkpoint at each step as the original models for p-diff training. For our hand-designed architectures, we also first train the models from scratch and then follow the same fine-tuning procedures. Note that during the fine-tuning procedures, we fine-tune all the model parameters rather than a subset of the model.
The noise scales ($\sigma_v$ and $\sigma_z$) for the Gaussian noise added to the parameters and latent representations are set to 0.001 and 0.1, respectively.
In most cases, data collection and training for p-diff can be completed within 1 to 3 hours using a single NVIDIA A100 GPU with 40GB of memory.

\paragraph{Inference details.}
We synthesize 200 sets of novel parameters by feeding random noise into the trained diffusion model and decoder. These synthesized parameters are then concatenated with the aforementioned remaining parameters to form the generated models. By default, p-diff is compared to the baselines using \emph{the best validation accuracy}.

\paragraph{Baselines.}
i) The best validation accuracy of the original models is denoted as `original'. ii) Average weight ensemble~\cite{krogh1994neural, wortsman2022model} of the original models is denoted as `ensemble'.

\subsection{Results}
\label{sec:results}
\paragraph{Performance on partial parameter generation.}
Table~\ref{tab:main} compares the performance of the generated models with the two baselines across 7 datasets and 6 architectures.
We make several notable observations: 
i) In most cases, our method achieves similar or better results compared to both baselines. This demonstrates that our method can efficiently learn the distribution of high-performing parameters and generate superior models from random noise.
ii) Our method consistently performs well across all datasets and architectures, highlighting its strong generality.

\paragraph{Generalization to full parameter generation.}
Until now, we have evaluated the effectiveness of our approach in synthesizing a subset of model parameters, \textit{i.e.,} batch normalization parameters. \textit{What about synthesizing entire model parameters?} We validate our approach's capability to generate full neural network parameters across 5 small, hand-designed architectures (ranging from 24K to 81K parameters due to GPU memory constraints): ConvNet-mini, MLP-mini, ResNet-mini, ViT-mini, and ConvNeXt-mini.

We evaluate full parameter synthesis with the five hand-designed architectures on CIFAR-10 and STL-10 datasets. The results are presented in Table~\ref{tab:all_para}.
It is worth noting that all architectures include unique structures. For example, ResNet-mini includes skip connections, ViT-mini contains self-attention mechanisms, and ConvNeXt-mini has depthwise separable convolutions.
The results demonstrate that our synthesized models achieve comparable or even better performance than the original ones across all architectures.

\vspace{-.5em}
\begin{table*}[htp]
\label{tab:ablations}
\centering
\subfloat[\textbf{Noise augmentation.} Using noise augmentation is important for p-diff to generate model parameters with high performance.
\label{tab:abl_of_noise_augmentation}]{
\centering
\begin{minipage}[h]{0.29\textwidth}
\begin{center}
\centering
\tablestyle{6pt}{1.15}
\begin{tabular}{lcc}
    noise augmentation & best  & average \\
    \shline
    none & 77.5 & 76.7 \\
    parameter & 77.4 & 76.7 \\
    latent & 77.9 & 77.6 \\
    both & \baseline{77.9}
    & \baseline{\textbf{77.7}} \\
    &&
\end{tabular}
\end{center}
\end{minipage}
}
\hspace{2em}
\subfloat[\textbf{Parameter Noise.} Generated models' performance is less sensitive to parameter noise.
\label{tab:param_noise}]{
\centering
\begin{minipage}[h]{0.29\textwidth}
\begin{center}
\tablestyle{6pt}{1.15}
\begin{tabular}{lccc}
noise scale & best &  average \\
\shline
0.001 &  \baseline{77.9} & \textbf{\baseline{77.7}} \\
$\times$ 0.01 & 77.9 & 77.6\\
$\times$ 5 & 77.9 & 77.6 \\
$\times$ 2000 & 77.9 & 77.6 \\
$\times$ 5000 & 77.3 & 76.8 \\
\end{tabular}
\end{center}
\end{minipage}}
\hspace{2em}
\subfloat[\textbf{Latent noise.} Generated models' performance is more sensitive to latent noise.
\label{tab:latent_noise}]{
\centering
\begin{minipage}[h]{0.29\textwidth}
    \begin{center}
        \tablestyle{6pt}{1.15}
\begin{tabular}{lccc}
noise scale & best &  average \\
\shline
0.1 & \baseline{77.9} & \textbf{\baseline{77.7}} \\
$\times$ 0.01 & 77.4 & 76.7 \\
$\times$ 0.1 & 77.4 & 76.8 \\\
$\times$ 10 & 77.9 & 77.6 \\
$\times$ 100 & 77.5 & 77.2 \\
\end{tabular}
    \end{center}
\end{minipage}
}\\
\centering
\subfloat[\textbf{Optimizer.}
For original models trained with different optimizers, p-diff can generate similar or better performing parameters.
\label{tab:optimizer}]{
\centering
\begin{minipage}[h]{0.29\textwidth}
    \begin{center}
\tablestyle{4pt}{1.15}
\begin{tabular}{lccc}
optimizer  & original & best & average  \\
\shline
SGD & 77.6 & 77.5 & 77.3 \\
Adam~\cite{kingma2014adam} & 79.7 & 79.6 & 79.4\\
AdamW~\cite{loshchilov2017decoupled} &\baseline{77.4}&  \baseline{77.9} &\baseline{77.7} \\
\\
\\
\end{tabular}
    \end{center}
\end{minipage}}
\hspace{2em}
\subfloat[
\textbf{Number of pretrained model samples.}
A larger $K$ can improve the performance of generated model parameters.
\label{tab:K_ablation}]{
\centering
\begin{minipage}[h]{0.29\textwidth}
\begin{center}
\tablestyle{6pt}{1.15}
\begin{tabular}{lccc}
    samples ($K$) & best & average \\
    \shline
    10 & 77.0 & 76.6 \\
    50 & 77.5 & 77.2 \\
    200 & 77.8 & 77.5 \\
    300 & \baseline{77.9} & \baseline{\textbf{77.7}} \\
    400 & \textbf{78.0} & 77.6 \\
\end{tabular}
\end{center}

\end{minipage}}
\hspace{2em}
\subfloat[\textbf{Subsets of model parameters}. P-diff consistently generates similar or better performing parameters across parameter groups.
\label{tab:para_type}]{
\centering
\begin{minipage}[h]{0.29\textwidth}
\begin{center}
    
\tablestyle{3pt}{1.15}
\begin{tabular}{lccc}
layer & orginal & best & average \\
\shline
conv. layer (1)& 77.8 &77.8& 77.3\\
fc layer&77.8 & 77.8 & 77.7\\
BN layers (10-13)& 77.4 & 77.5 & 77.0 \\
BN layers (14-15)& 77.7 & 77.3 & 76.8\\
BN layers (16-17)& \baseline{77.4} & \baseline{77.9}& \baseline{77.7} 
\end{tabular}
\end{center}
\end{minipage}}
\vspace{-.5em}
\caption{\textbf{P-diff ablation experiments.} We present the best validation accuracy of the original models and the best and the average accuracies of the generated models. Unless stated otherwise, p-diff is trained on $K=300$ samples of BN parameters at the 16th and 17th layers of ResNet-18, with parameter and latent noise augmentation applied. Defaults are marked in \colorbox{baselinecolor}{gray}.
\textbf{Bold entries} are best results.}
\end{table*}

\subsection{Ablation Analysis}
\label{sec:main_abl}
In this section, we conduct extensive ablation studies to demonstrate our method's characteristics in different data collection and training settings. We focus our analysis on ResNet-18 and CIFAR-100, and report both the best accuracy and the average accuracy across all generated models.

\paragraph{Noise augmentation.} In Section~\ref{sec:param_autoencoder}, we propose
noise augmentation during training to enhance the robustness and generalization of the parameter autoencoder. Here, we ablate the effectiveness of applying the augmentations to the input parameters and latent representations, respectively. 
Results in Table~\ref{tab:abl_of_noise_augmentation} show that:
i) Noise augmentation is important for generating high-performing neural network parameters.
ii) Applying noise augmentation to the latent representations yields greater performance gains than applying it to the input parameters.
iii) Our default setting, which jointly applies noise augmentation to both input parameters and latent representations, achieves the best results.

We further evaluate the sensitivity of model performance to the scale of the noise applied to input parameters and latent representations. To do this, we vary one noise scale ($\sigma_v$ or $\sigma_z$) while keeping the other constant. The resulting accuracies in Table~\ref{tab:param_noise} and Table~\ref{tab:latent_noise} show that generated models are more sensitive to latent noise. Also, the current set of noise scales achieves the optimal performance.

\paragraph{Generalization to other optimizers.}
Our default setting employs the AdamW~\cite{loshchilov2017decoupled} optimizer for training the original models. To validate our method's robustness across different optimization algorithms, we conduct experiments using SGD and Adam~\cite{kingma2014adam}, with other experimental protocols detailed in the previous section remain the same. The results in Table~\ref{tab:optimizer} demonstrate that our approach achieves comparable performance to the original models across all three optimizers, confirming its generality.

\paragraph{The number of training models.} Table~\ref{tab:K_ablation} varies the size of p-diff's training data ($K$), \textit{i.e.}, the number of saved original model checkpoints. 
We observe that both the best accuracy and the average accuracy consistently improve as the number of original models increases, gradually converging as $K$ reaches 400. This indicates the feasibility of our method across different training data sizes.
We note that the model performance with a small number of original models ($K$ = 10) is lower than those with larger sizes of training data. This is expected, since it may be very difficult for the diffusion model to learn the target distribution effectively when only a few samples are provided during training.

\paragraph{Where to apply p-diff.}
We default to synthesizing the parameters of the last two batch normalization layers in ResNet-18.
To investigate the robustness of our approach in generating other parameters, we also conduct experiments on batch normalization layers at different depths, as well as other types of layers. As shown in Table~\ref{tab:para_type}, 
our approach consistently generates models that perform on par with the original ones across BN layers at all depths.
Additionally, generated parameters on both convolutional and fully-connected layers in our approach achieve reasonable performance compared with the ones in original models.

\vspace{-.5em}
\begin{figure*}[h]
    \centering
    \subfloat[epoch = 10 (default)]{
    \centering
    \begin{minipage}[h]{0.328\linewidth}
    \begin{center}
        
        \includegraphics[width=\linewidth]{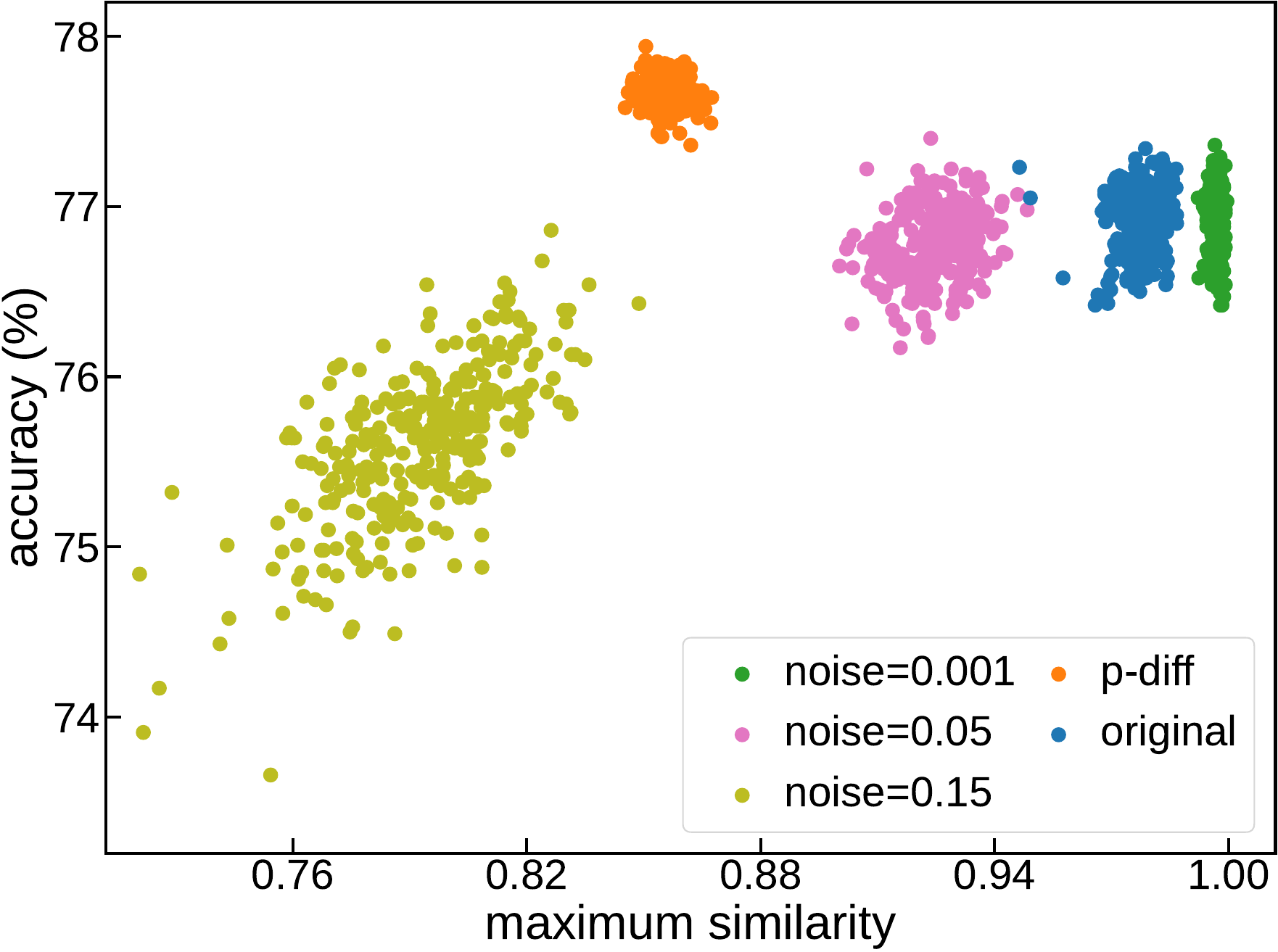}
    \end{center}
    \end{minipage}}
    \hfill
    \centering
    \subfloat[epoch = 3]{
    \centering
    \begin{minipage}[h]{0.328\linewidth}
    \begin{center}        
        \includegraphics[width=\linewidth]{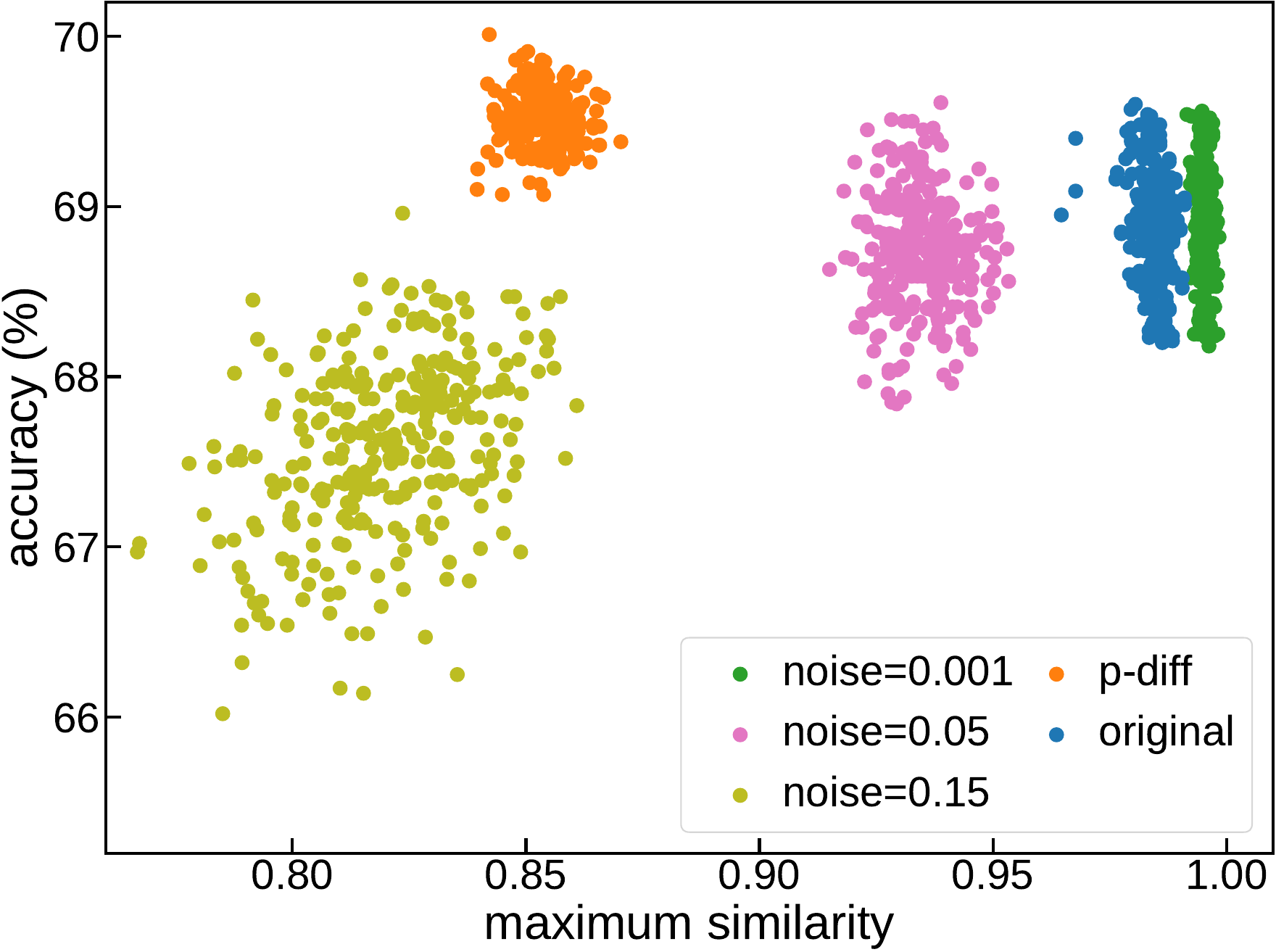}
    \end{center}
    \end{minipage}}
    \hfill
    \centering
    \subfloat[epoch = 1]{
    \centering
    \begin{minipage}[h]{0.328\linewidth}
    \begin{center}
        \includegraphics[width=.99\linewidth]{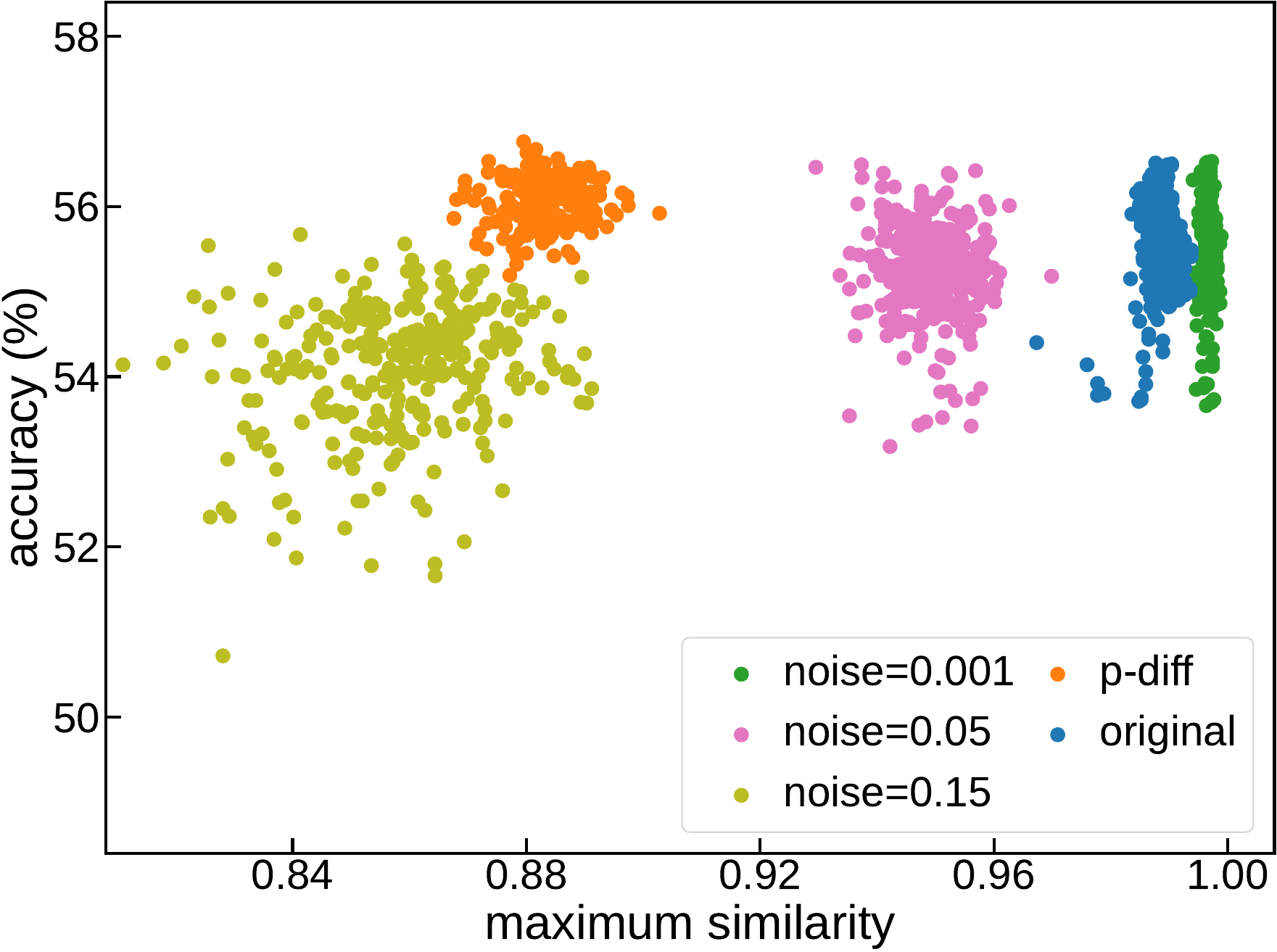}
    \end{center}
    \end{minipage}}
\vspace{-.5em}
    \caption{\textbf{p-diff generalizes to under-trained original models.} For original models trained for 1 or 3 epochs before the fine-tuning steps, the generated models can still achieve high accuracy with low similarity to original models. Results are on ResNet-18 and CIFAR-100.}
    \label{fig:diff_epoch}
\end{figure*}

\begin{figure*}[h]
    \centering
    \subfloat[learning rate = 0.3
    \label{fig:3a}]{
    \centering
    \begin{minipage}[h]{0.328\linewidth}
    \begin{center}
        
        \includegraphics[width=\linewidth]{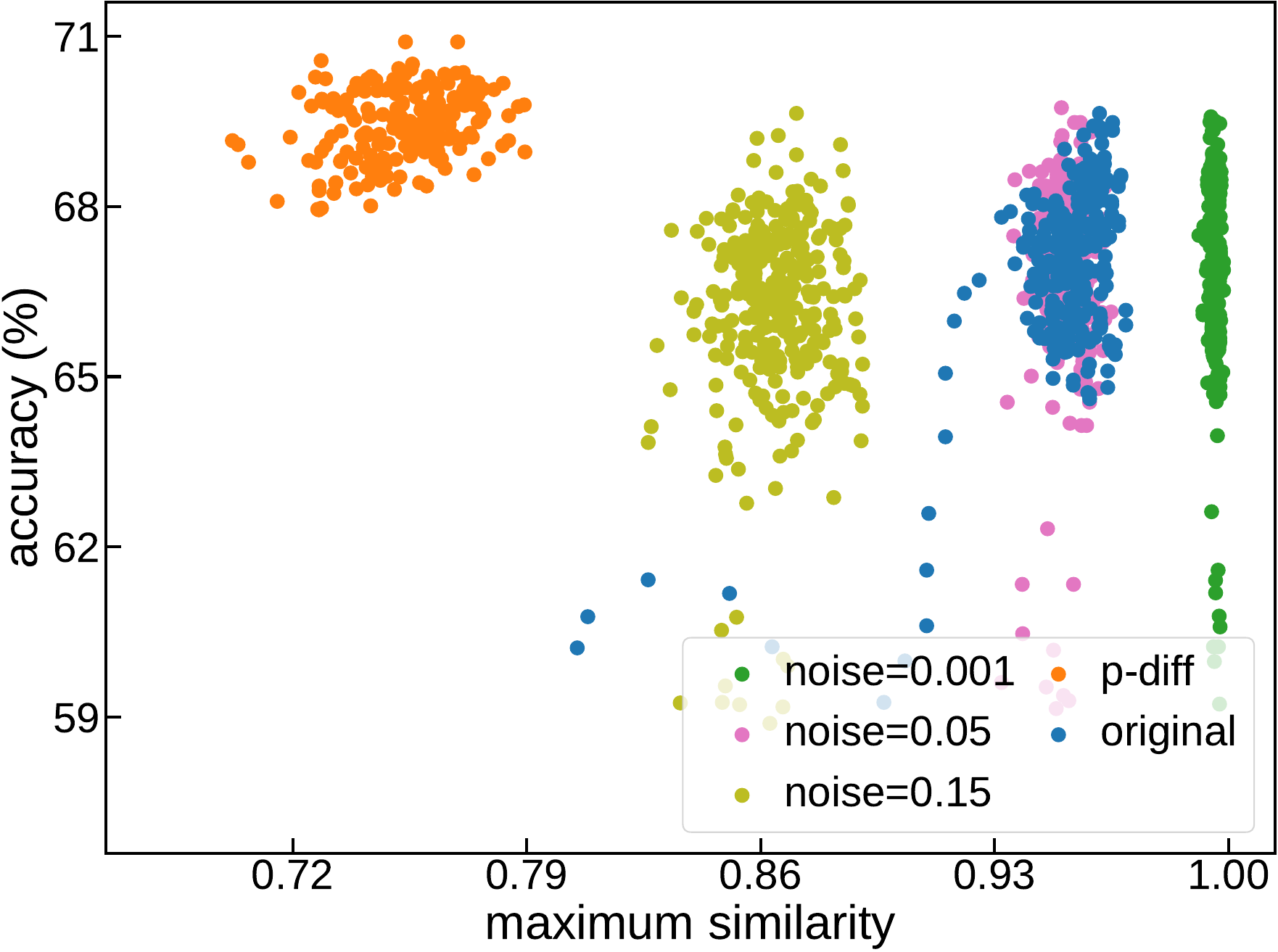}
    \end{center}
    \end{minipage}}
    \hfill
    \centering
    \subfloat[learning rate = 0.03 (default)]{
    \centering
    \begin{minipage}[h]{0.328\linewidth}
    \begin{center}
        \includegraphics[width=.99\linewidth]{figures/baseline.pdf}
    \end{center}
    \end{minipage}}
    \hfill
    \centering
    \subfloat[learning rate = 0.0003]{
    \centering
    \begin{minipage}[h]{0.328\linewidth}
    \begin{center}        
        \includegraphics[width=\linewidth]{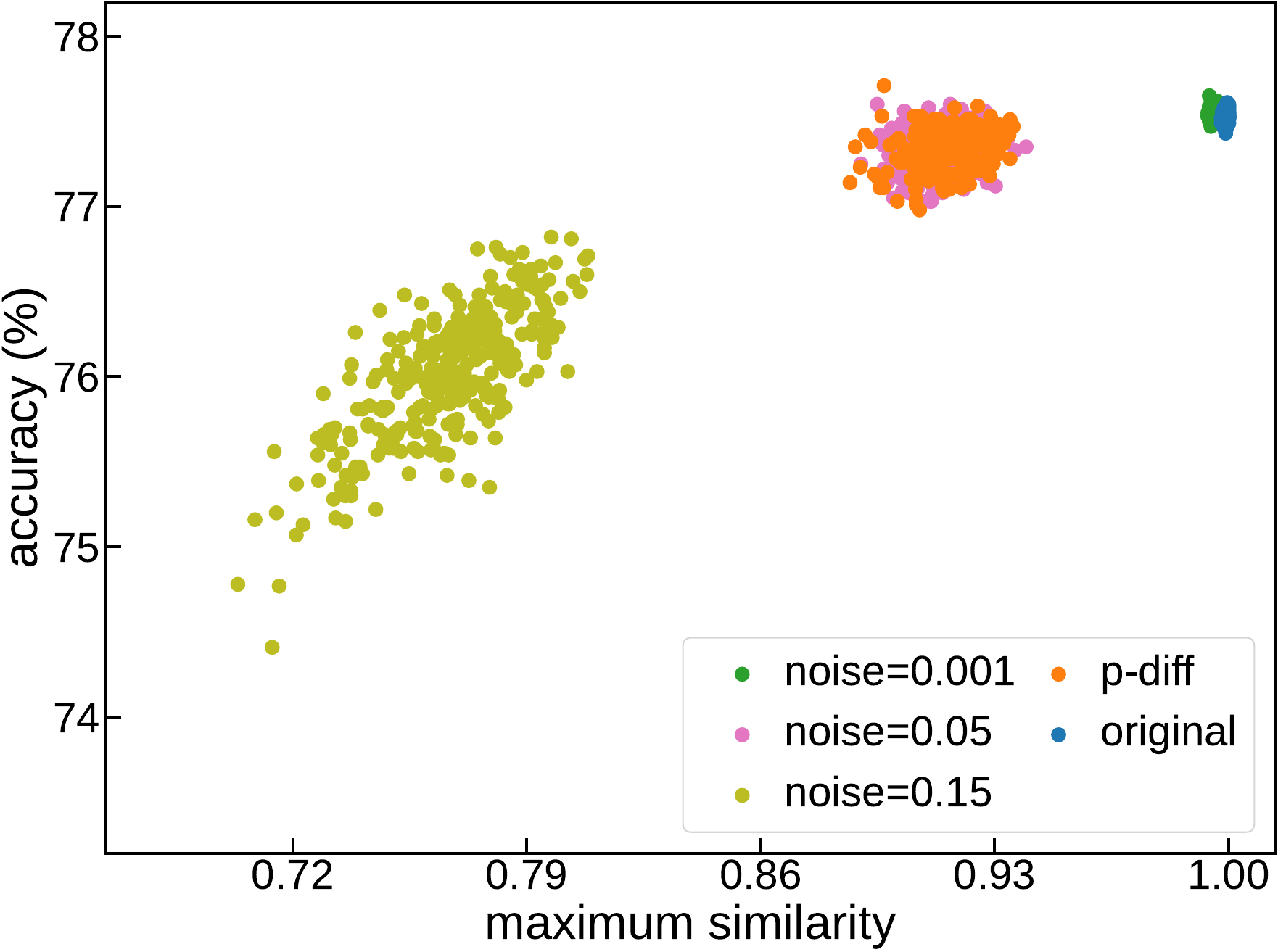}
    \end{center}
    \end{minipage}}
\vspace{-.5em}
    \caption{\textbf{Diversity of original models is important for the novelty of generated models.} We use learning rates of 0.03, 0.003, and 0.3 for fine-tuning and saving the normalization layers of the converged model as training samples. Higher learning rate leads to greater original model diversity and lower maximum similarity of generated models. Results are on ResNet-18 and the CIFAR-100 dataset.}
    \label{fig:diff_lr}
\vspace{-.25em}
\end{figure*}
\section{Is P-diff Only Memorizing?}
We have demonstrated above that p-diff is able to generate models with high performance. Nevertheless, this is meaningful only if these generated models are quite different from the training ones, while still capturing the distribution of high-performing model checkpoints.

In this section, we investigate the difference among the original, noise-added, and generated models. 
We propose a similarity metric for model checkpoints to quantitatively assess the novelty of models generated by p-diff.

\paragraph{Questions and experiment designs.} Here, we first ask the following questions: 1) Does p-diff merely memorize the original model checkpoints in the training set? 2) Are generated models equivalent to the original models with Gaussian noise added? For p-diff to truly capture the distribution of high-performing checkpoints, it should generate novel parameters that perform differently from the original models. To verify this, we examine and visualize the differences in model predictions among original, noise-added, and generated checkpoints. We carry out our analysis on ResNet-18~\citep{he2016deep} and CIFAR-100~\citep{krizhevsky2009learning} under the default setting.

\paragraph{Similarity metric.}
While Euclidean and cosine distances provide parameter-level comparisons, they cannot account for the difference in parameter value range across parameter groups. Most importantly, they fail to capture the actual behavioral differences between models. Therefore, we measure the similarity between a pair of models by calculating the Intersection over Union (IoU) of their wrong predictions. The IoU is formulated as follows,
\begin{equation}
    \mathrm{IoU} = |P_{1}^{\mathrm{wrong}} \cap P_{2}^{\mathrm{wrong}}|/|P_{1}^{\mathrm{wrong}} \cup P_{2}^{\mathrm{wrong}}|,
    \label{eqn:iou}
\end{equation}
where $P_{\cdot}^{\mathrm{wrong}}$ denotes the indices of wrong predictions on the validation set, $\cap$ and $\cup$ represent union and intersection operations of sets. A lower IoU indicates a lower similarity between the predictions of the two models and is desired. From now on, we use the IoU of wrong predictions as the similarity metric throughout our paper.

\paragraph{Measure novelty with maximum similarity.}
For each model checkpoint, we calculate its similarity to every original model and take the maximum similarity as a measure of the model checkpoint's novelty.
We compute the maximum similarity of each original, noise-added, and generated model. Note that when evaluating the maximum similarity of an original model, we exclude its similarity to itself and only consider the similarity to other original models.

To compare p-diff generated models with Gaussian noise-added models, we evaluate the latter at noise scales of 0.001, 0.05, and 0.15. In Figure~\ref{fig:3a}, we plot validation accuracy against maximum similarity for the original (\textcolor{blue}{blue}), noise-added, and p-diff generated models (\textcolor{orange}{orange}). As the noise scale increases, the accuracy and maximum similarity of the noise-added models decrease simultaneously. In contrast, p-diff's generated models are all located in the upper-left region, indicating both high accuracy and low maximum similarity. This suggests that our approach achieves an effective trade-off between accuracy and novelty.

\vspace{-.5em}
\begin{figure*}[tp]
    \centering
    \subfloat[noise augmentations.
    \label{fig:5a}]{
    \centering
    \begin{minipage}[h]{0.328\textwidth}
    \begin{center}        
        \includegraphics[width=\textwidth]{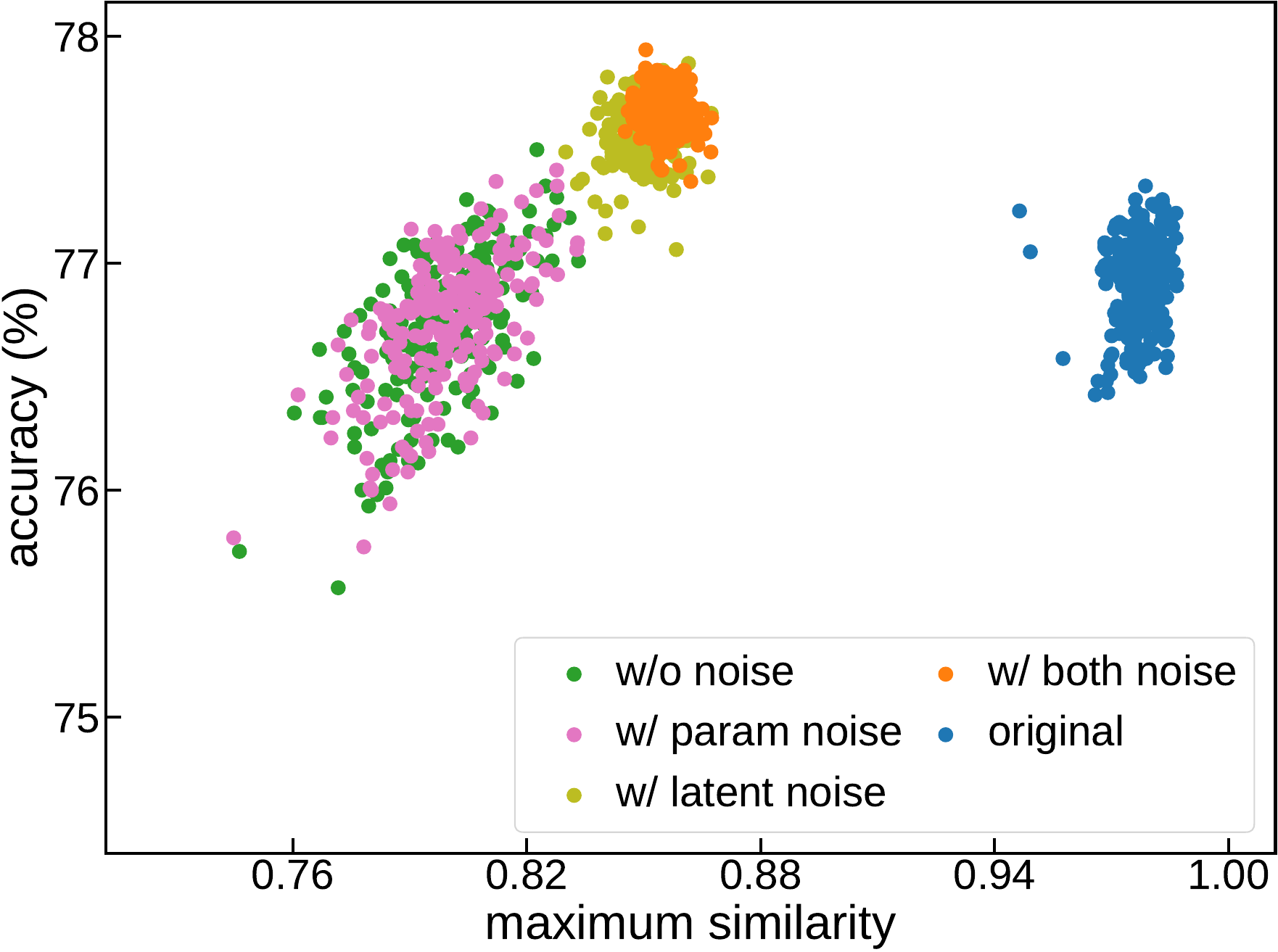}
    \end{center}
    \end{minipage}}
    \hfill
    \centering
    \subfloat[accuracy vs diffusion steps.
    \label{fig:5b}]{
    \centering
    \begin{minipage}[h]{0.328\textwidth}
    \begin{center}        
        \includegraphics[width=\textwidth]{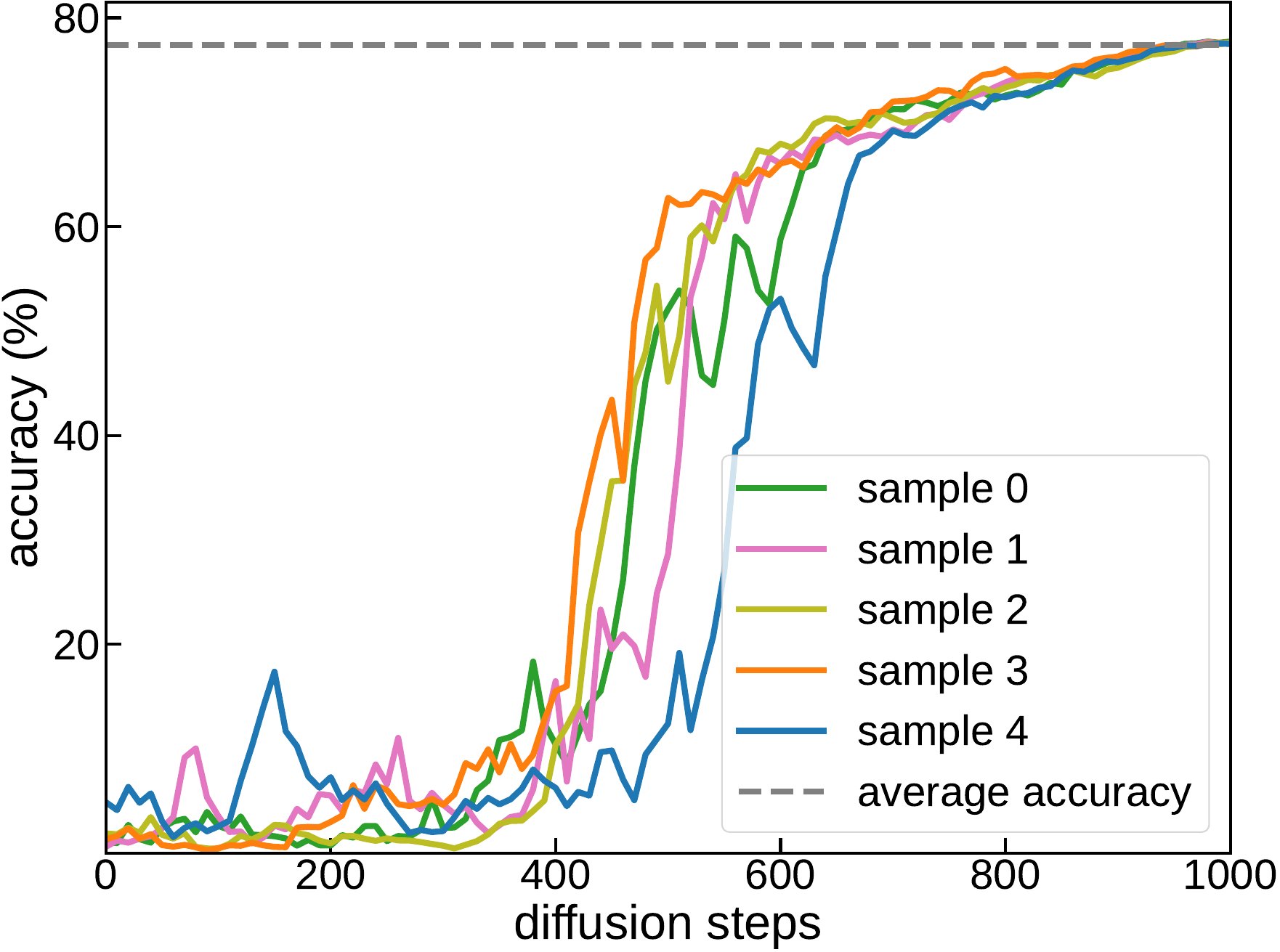}
    \end{center}
    \end{minipage}}
    \hfill
    \centering
    \subfloat[number of training samples.\label{fig:5c}]{
    \centering
    \begin{minipage}[h]{0.328\textwidth}
    \begin{center}        
        \includegraphics[width=\textwidth]{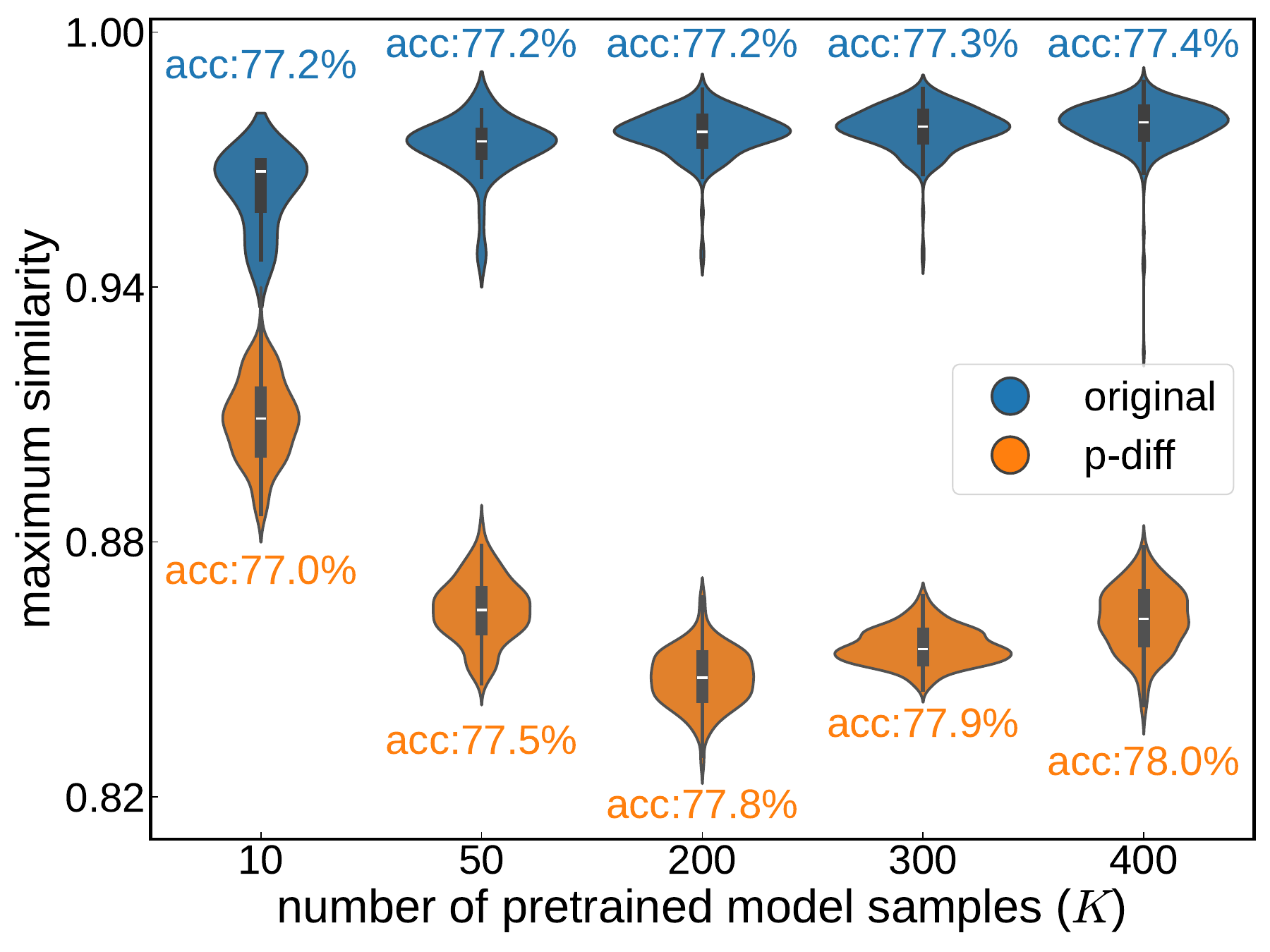}
    \end{center}
    \end{minipage}}
    \caption{(a) shows the impact of parameter and latent noise augmentation on the novelty and accuracy of generated models. Using both parameter and latent noise augmentation achieves the highest accuracy. (b) illustrates the accuracy trajectories across different diffusion steps during inference. Our approach generates high-performing parameters through diverse paths. (c) presents the distributions of maximum similarity, where thickness indicates density. A large set of original models leads to higher novelty in generated models.}
\end{figure*}

\paragraph{Under-trained original models.} In the default setting, we use a well-trained model and save its checkpoints from 300 additional fine-tuning steps as training samples for p-diff. Here, we explore whether a model trained for a shorter duration can still enable p-diff to generate novel parameters with reasonable performance compared to the original ones.

Concretely, we train models for only 1 or 3 epochs, fine-tune the selected subset of parameters, and then save checkpoints. The accuracy and maximum similarity of the original, generated, and noise-added models are plotted in Figure~\ref{fig:diff_epoch}. As expected, the accuracy of these original models is much lower due to shorter training. However, in all cases, p-diff can still achieve better trade-off between accuracy and similarity than noise addition to the original models.

\paragraph{Original model diversity.}
p-diff is designed to capture the underlying distribution of trained high-performing neural network parameters. Therefore, it is important to understand how the diversity of the original models impact the performance and novelty of the generated models. We control the diversity of original models by varying the learning rate during the fine-tuning steps, where we save the original models. Specifically, we experiment with a smaller learning rate (0.0003) and a larger learning rate (0.3), in addition to our default rate (0.03). The results are presented in Figure~\ref{fig:diff_lr}.

We observe that the diversity among original models, measured by maximum similarity, indeed increases with higher fine-tuning learning rate. When training p-diff on model checkpoints fine-tuned with a small learning rate (0.0003), the generated models from p-diff are very similar to the original ones. More importantly, they exhibit a trade-off between validation accuracy and maximum similarity that is similar to that of simple noise addition. However, as the fine-tuning learning rate increases and the diversity of original models grows, p-diff, trained on these more diverse models, shows a clear advantage over noise addition. This highlights the importance of original model's diversity in enhancing the novelty of p-diff's generated models.

\paragraph{Impact of noise augmentation.}
While Table~\ref{tab:abl_of_noise_augmentation} examines noise augmentation's impact on accuracy, here we take a closer look at its impact on generated model's similarity to original models. 
In Figure~\ref{fig:5a}, we plot the accuracy against maximum similarity of the generated models when one or both of the noise augmentations are not used. Augmentation with both latent and input parameter noise enhances model performance while decreasing the maximum similarity.

Additionally, we vary the noise scale and evaluate its impact on the maximum similarity range and best accuracy of p-diff generated models. The results show that our default noise augmentation scale obtains the best trade-off between novelty and accuracy. A small noise augmentation scale reduces the robustness of the parameter autoencoder, resulting in degraded accuracy. Conversely, while larger noise augmentation scales can improve model performance, they negatively impact the diversity of the generated models.

\vspace{-.5em}
\begin{table}[h]
\centering
\tablestyle{5pt}{1.2}
\begin{tabular}{lx{35}c}
para \& latent noise scales & best & similarity range  \\
\shline
0.001 \& 0.1 (default) & 77.9 & 0.85 $\sim$ 0.87 \\
\hline
$\times$ 0.01 & 77.4 & 0.75 $\sim$ 0.84 \\
$\times$ 0.1 & 77.4 & 0.75 $\sim$ 0.83 \\
$\times$ 10 & 77.9 & 0.82 $\sim$ 0.87\\
$\times$ 100 & 77.6 & 0.79 $\sim$ 0.83
\end{tabular}
\caption{\textbf{Different noise scales.} Larger noise scales can lead to higher accuracy and higher maximum similarity range.}
\vspace{-.75em}
\end{table}

\paragraph{Parameter trajectories of the p-diff process.}
To investigate the effectiveness of our method, p-diff, we visualize the accuracy trajectories across different timesteps during the inference stage.
Specifically, we first sample five different random noises as the inputs to the diffusion model, and then save the intermediate latent vectors. After that, the intermediate latent vectors are fed into the trained decoder to synthesize the models. Finally, we evaluate the generated models and visualize their accuracy with respect to the number of diffusion steps in Figure~\ref{fig:5b}.

All trajectories eventually converge to accuracy higher than the average original model accuracy of 77.4\%. Importantly, The diverse accuracy trajectories across different random seeds suggest that our method can synthesize model parameters through distinct paths in the parameter space.

\paragraph{From memorization to novelty.}
A diffusion model can better capture the underlying distribution of training data when a larger number of data points are sampled from that distribution.
To examine how the number of original models ($K$) affects the novelty of generated models in p-diff, we vary $K$ and visualize the distribution of maximum similarities of all original and generated models with respect to the original models in Figure~\ref{fig:5c}. Additionally, we report the mean accuracy for both original and generated models.

Across all values of $K$, p-diff's generated models maintain comparable performance to the original models. When $K$ = 10, the generated models exhibit high similarities, all concentrated in a narrow range. With a lower $K$, p-diff is primarily memorizing the original models. However, as $K$ increases, the maximum similarity decreases, suggesting that p-diff benefits from the increased training checkpoint diversity to generate distinct yet effective parameters. The similarity and accuracy stabilize at $K$ = 300 and improve marginally when p-diff is provided with 400 checkpoints.

\section{Related Work}
\label{RW}

\paragraph{Diffusion models.} Diffusion models have achieved remarkable success in visual generation. 
These methods~\cite{ho2020denoising, dhariwal2021diffusion, ho2022imagen, Peebles2022DiT, hertz2023prompttoprompt, li2023your} are based on non-equilibrium thermodynamics~\cite{jarzynski1997equilibrium, sohl2015deep}. Their development trajectory is similar to GANs~\cite{zhu2017unpaired, isola2017image, brock2018large}, VAEs~\cite{kingma2013auto, razavi2019generating}, and flow-based models~\cite{dinh2014nice, rezende2015variational}, expanding from unconditional generation to generation conditioned on text, images, or other structured information. Research on diffusion models can be categorized into three main branches. The first branch focuses on enhancing the synthesis quality of diffusion models, such as DALL$\cdot$E 2~\cite{ramesh2022hierarchical}, Imagen~\cite{saharia2022photorealistic}, and Stable Diffusion~\cite{rombach2022high}. The second branch aims to improve the sampling speed, 
including DDIM~\cite{song2021denoising}, Analytic-DPM~\cite{bao2022analyticdpm}, and DPM-Solver~\cite{lu2022dpmsolver}. The final branch involves reevaluating diffusion models from a continuous perspective~\cite{song2019generative, feng2023score}.

\paragraph{Parameter generation.} 
HyperNet~\cite{ha2017hypernetworks} dynamically generates the weights of a model with variable architecture. SMASH~\cite{brock2018smash} introduces a flexible scheme based on memory read-writes that can define a diverse range of architectures. G.pt~\cite{peebles2023learning} trains a transformer-based diffusion model conditioned on current parameters and target loss or error to generate new parameters. However, the generated models tend to underperform the training model checkpoints.
HyperRepresentations~\cite{schurholt2022hyper, schurholt2022hyper, schurholt2024towards} uses an autoencoder to capture the latent distribution of trained models. These works focus primarily on the analysis of the parameter space rather than generating high-performing parameters.

\paragraph{Parameter space learning.}
MetaDiff~\cite{zhang2023metadiff} introduces a diffusion-based meta-learning method for few-shot learning, where a layer is replaced by a diffusion U-Net~\cite{ronneberger2015u}. Diffusion-SDF~\cite{chou2023diffusion} proposes a diffusion model for shape completion, single-view reconstruction, and reconstruction of real-scanned point clouds.
HyperDiffusion~\cite{erkocc2023hyperdiffusion} uses a diffusion model on MLPs to generate new neural implicit fields \textcolor{black}{(a representation for 3D shape)}.
\textcolor{black}{DWSNets~\cite{navon2023equivariant} introduces a symmetry-based approach for designing neural architectures that operate in deep weight spaces.}
\textcolor{black}{NeRN~\cite{ashkenazi2023nern} shows that  neural representations can be used to directly represent the weights of a pre-trained convolutional neural network.}
Different from them, our method is designed to learn high-performing parameter distributions.

\paragraph{Stochastic and Bayesian neural networks.}
Our approach could be viewed as learning a prior over network parameters, represented by the trained diffusion model. Learning parameter priors for neural networks has been studied in classical literature. 
Stochastic neural networks (SNNs)~\cite{sompolinsky1988chaos, wong1991stochastic, schmidt1992feed,murata1994network} also learn such priors by introducing randomness to improve the robustness and generalization of neural networks.
The Bayesian neural networks~\cite{neal2012bayesian, kingma2013auto, rezende2014stochastic, kingma2015variational, gal2016dropout} aim to model a probability distribution over neural networks to mitigate overfitting, learn from small datasets, and estimate the uncertainty of model predictions.
\cite{graves2011practical} proposes an easily implementable stochastic variational method as a practical approximation to Bayesian inference for neural networks. They introduce a heuristic pruner to reduce the number of network weights, resulting in improved generalization. 
\cite{welling2011bayesian} combines Langevin dynamics with SGD to incorporate a Gaussian prior into the gradient. This transforms SGD optimization into a sampling process. 
\textit{Bayes by Backprop}~\cite{blundell2015weight} learns a probability distribution prior over the weights of a neural network.  
These methods mostly operate in small-scale settings, while p-diff shows its effectiveness and potential in real-world neural network architectures.

\section{Discussion and Conclusion}
Neural networks have several conventional learning paradigms, such as supervised learning~\citep{krizhevsky2012imagenet, simonyan2014very, he2016deep, dosovitskiy2020image} and self-supervised learning~\citep{devlin2018bert, brown2020language, he2020momentum, he2022masked}. In this study, we demonstrate that diffusion models can also be employed to generate high-performing neural network parameters across tasks, network architectures, and training settings. Notably, we demonstrate that our method can generate novel models with distinct behaviors compared to the training checkpoints. Using diffusion process for neural network parameter updates is a potential novel paradigm in deep learning.

We acknowledge that images / videos and parameters are signals of different natures, and this distinction must be handled with care. While diffusion models excel in image / video generation, their applications to model parameters remain relatively underexplored. This gap presents a series of challenges for neural network parameter diffusion. Our approach marks an initial step toward address some of these challenges. Nevertheless, there are still unresolved challenges, including memory constraints for generating the full parameters of large architectures, the efficiency of structure designs, and the stability of performance.

\paragraph{Acknowledgments.} We thank Kaiming He, Dianbo Liu, Mingjia Shi, Zheng Zhu, Bo Zhao, Jiawei Liu, Yong Liu, Ziheng Qin, Zangwei Zheng, Yifan Zhang, Xiangyu Peng, Hongyan Chang, Dave Zhenyu Chen, Ahmad Sajedi, and George Cazenavette for valuable discussions and feedbacks. This research is supported by the National Research Foundation, Singapore under its AI Singapore Programme (AISG Award No: AISG2-PhD-2021-08-008).

{
    \small
    \bibliographystyle{ieeenat_fullname}
    \bibliography{main}
}

\clearpage

\appendix
\section{Experimental Settings}
In this section, we detail the experimental settings and provide instructions for reproduction.
\subsection{Training Recipe}
We provide our basic recipes to train p-diff for ResNet-18 and CIFAR-100 in Table~\ref{tab:exp_conf}. For other datasets, the learning rate and training iterations need to be slightly adjusted.

\begin{table}[h]
\begin{subtable}[t]{\linewidth}
\tablestyle{6pt}{1.02}
\centering
\begin{tabular}{y{96}|x{98}}
config & value \\
\shline
optimizer & AdamW \\
learning rate & 5e-4 \\
weight decay & 5e-4 \\
training epochs & 10 \\
optimizer momentum & 0.9, 0.999 \\
batch size & 128 \\
learning rate schedule & cosine decay \\
augmentation & \texttt{RandomResizedCrop}~\cite{szegedy2015going}\\
\end{tabular}
\caption{training recipe for model checkpoints}
\end{subtable}
\vspace{1.5em}

\begin{subtable}[t]{\linewidth}
\centering
\tablestyle{6pt}{1.02}
\begin{tabular}{y{96}|x{98}}
config & value \\
\shline
optimizer & AdamW \\
learning rate & 3e-2 \\
weight decay & 5e-4 \\
finetuning iterations & 200 \\
optimizer momentum & 0.9, 0.999 \\
batch size & 128 \\
learning rate schedule & cosine decay \\
augmentation & \texttt{RandomResizedCrop}~\cite{szegedy2015going}\\
\end{tabular}
\caption{finetuning recipe for model checkpoints}
\end{subtable}
\vspace{1.5em}

\begin{subtable}[t]{\linewidth}
\centering
\tablestyle{6pt}{1.02}
\begin{tabular}{y{96}|x{98}}
config & value \\
\shline
optimizer & AdamW \\
learning rate & 2e-5 \\
weight decay & 0 \\
training iterations & 500 \\
optimizer momentum & 0.9, 0.999 \\
batch size & 50 \\
learning rate schedule & cosine decay \\
noise augmentation & $\sigma_v,\sigma_z{=}$1e-3, 1e-1\\
\end{tabular}
\caption{training recipe for parameter autoencoder}
\end{subtable}
\vspace{1.5em}

\begin{subtable}[t]{\linewidth}
\centering
\tablestyle{6pt}{1.02}
\begin{tabular}{y{96}|x{98}}
config & value \\
\shline
optimizer & AdamW \\
learning rate & 1e-4 \\
weight decay & 0 \\
training iterations & 1000 \\
optimizer momentum & 0.9, 0.999 \\
batch size & 50 \\
learning rate schedule & cosine decay \\
diffusion steps & $T{=}$1000 \\
noise variance & $\beta_1,\beta_T{=}$1e-4, 2e-2 \\
noise schedule & linear
\end{tabular}
\caption{training recipe for diffusion model}
\end{subtable}
\caption{\textbf{Training recipes.}}
\label{tab:exp_conf}
\end{table}

\subsection{Hand-designed Architectures}

In Section~\ref{sec:results}, we evaluate our method's ability to generate full model parameters by applying it to five small hand-designed architectures: ConvNet-mini, MLP-mini, ResNet-mini, ViT-mini, and ConvNeXt-mini. Here, we use CIFAR-10 as an example and show the details of these architectures.

\emph{ConvNet-mini.} conv1: (in\_channels = 3, out\_channels = 8, kernel\_size = 7),  conv2: (8, 8, 3),  conv3: (8, 4, 3), linear1: (in\_features = 64, out\_features = 16), linear2: (16, 10).

\emph{MLP-mini.} linear1: (1024, 64), linear2: (64, 16), linear3: (16, 10).

\emph{ResNet-mini.} conv1: (3, 8, 3), residual block1: (in\_chan
nels = 8, out\_channels = 8, kernel\_size = 3, bottleneck = 4), residual block2: (8, 8, 3, 4), linear1: (128, 16), linear2: (16, 10).

\emph{ViT-mini.} number tokens: 64, token dimension: 64, depth: 4, number heads: 4, head dimension: 8, feed forward dimension: 64.

\emph{ConvNeXt-mini.} stage depths: (1, 2, 2, 2), stage dimensions: (16, 32, 32, 32).

\subsection{Number of Generated Parameters}
In Table~\ref{tab:parameter_number}, we list the number of model parameters to generate for different architectures used in our experiments.

\begin{table}[h]
\centering
\tablestyle{6pt}{1.2}
\begin{tabular}{c c c c}

dataset & architecture & generation mode & parameter number \\
\shline
STL-10 & ResNet-18        & partial & 2048 \\
STL-10 & ResNet-50       & partial & 5120 \\
STL-10 & ViT-Tiny       & partial & 768 \\
STL-10 & ViT-Base       & partial & 3072 \\
STL-10 & ConvNeXt-Tiny       & partial & 3072 \\
STL-10 & ConvNeXt-Base      & partial & 4096 \\
STL-10 & ConvNet-mini     & entire & 24262 \\
STL-10 & MLP-mini     & entire & 66970 \\
STL-10 & ResNet-mini     & entire & 25138 \\
STL-10 & ViT-mini     & entire & 80714 \\
STL-10 & ConvNeXt-mini     & entire & 74970 \\
\end{tabular}
\caption{Number of parameters to generate for different architectures in our experiments. We take STL-10 as an example.}
\label{tab:parameter_number}
\end{table}

\section{Additional Experiments}

\subsection{Additional Baselines}
We also compare our approach with hyper-representations methods~\cite{schurholt2022hyper, schurholt2024towards} on MNIST and SVHN datasets. To ensure a fair comparison, we use p-diff to generate the same 3-layer convolutional network defined in the hyper-representations papers~\cite{schurholt2022hyper}.
As illustrated in Tab.~\ref{tab: hyper-representations}, p-diff achieves accuracy similar or better than the original accuracy, whereas models generated using the hyper-representations methods do not perform comparably to the original ones.

\begin{table}[h]
\centering
\tablestyle{4pt}{1.2}
\begin{tabular}{lccccc}
dataset & original & $S_{\mathrm{KDE30}}$ &SANE$_{\mathrm{SUB}}$ &SANE$_{\mathrm{KDE30}}$  & p-diff  \\
\shline
MNIST   & 92.1     & 68.6 &86.7&84.8                & \textbf{93.3}   \\
SVHN    & 71.3     & 54.5     &72.3&70.7            & \textbf{74.1}   \\
\end{tabular}
\caption{Model checkpoints generated by p-diff outperform those generated using the hyper-representation methods. Results are in \textit{average accuracy}.}
\label{tab: hyper-representations}
\end{table}

\subsection{Is Variational Autoencoder an Alternative to Our Approach?}
Variational autoencoder (VAE)~\cite{kingma2013auto}, as a powerful probabilistic generative model, has demonstrated remarkable success in various generation tasks. Here, we seek to use VAE to generate high-performing parameters. To ensure a fair comparison with p-diff, we implement a vanilla VAE using identical backbone architecture and training iterations.
As shown in Table~\ref{appendix:compare_vae_diff}, our proposed p-diff generates model parameters with much higher accuracies than VAE.

\begin{table}[h]
\centering
\tablestyle{5pt}{1.2}
\begin{tabular}{lx{35}x{35}c}
    method & best & average &similarity range\\
    \shline
    original &77.4 &76.9 &0.93$\sim$0.98 \\
    vae &77.3 &76.6 &0.74$\sim$0.82\\
    p-diff &\textbf{77.9} & \textbf{77.7} &0.85$\sim$0.87 \\
\end{tabular}
\caption{Comparisons between VAE and p-diff on ResNet-18 and CIFAR-100. Our approach achieves better performances.}
\label{appendix:compare_vae_diff}
\end{table}

\subsection{Diffusion Step}
We train the diffusion model by default with 1000 diffusion steps. Here, we conduct an ablation study to determine whether fewer steps can still achieve comparable results. As shown in Table~\ref{appendix:abl_of_time_step}, we observe that: i) Using too few timesteps can degrade the performance of the generated models. ii) Increasing the diffusion steps beyond 1000 does not provide significant improvement.

\begin{table}[h]
\centering
\tablestyle{5pt}{1.2}
\begin{tabular}{lx{35}x{35}c}
    diffusion step & best & average\\
    \shline
    10 & 74.6 & 70.5 \\
    100& 77.9 & 77.6 \\
    1000& 77.9 & \textbf{77.7} \\
    2000 & 77.8 & 77.5 \\
\end{tabular}
\caption{Effects of diffusion steps on ResNet-18 and CIFAR-100.}
\label{appendix:abl_of_time_step}
\end{table}

\subsection{P-diff on Other Vision Tasks}
So far, we have only demonstrated p-diff's ability to generate high-performing model weights for image recognition tasks. Below we test our method to generate models for other visual tasks (\textit{i.e.,} object detection, semantic segmentation, and image generation). For each task, we take pretrained models and finetune a subset of parameters in the models to collect 300 model checkpoints. These checkpoints are then used to train p-diff to generate new, high-performing model weights.

\paragraph{Object detection.}
For this task, we take the pretrained Faster R-CNN~\cite{ren2015faster} and FCOS~\cite{tian2020fcos} models, both of which use a ResNet-50 backbone and were trained on the COCO~\cite{lin2014microsoft} dataset. We report the results in Table \ref{tab:object_detection}. Our method generates models with similar or even better performance compared to the original ones. 

\begin{table}[htbp]
    \centering
    \tablestyle{10pt}{1.2}
    \begin{tabular}{y{50}x{20}x{35}}
        model & original & p-diff  \\
        \shline
         Faster R-CNN & 36.9 & \textbf{37.0} \\
         FCOS & 39.1 & 39.1 \\
    \end{tabular}
    \caption{P-diff on the object detection task. Results are reported in AP on the COCO dataset.}
    \label{tab:object_detection}
\end{table}

\paragraph{Semantic segmentation.}
Here we leverage FCN~\cite{long2015fully} with a ResNet-50 backbone to evaluate a subset of COCO val2017, on the 20 categories that are present in the Pascal VOC dataset. We generate the parameters of the last normalization layer of ResNet-50 and report the results in Table \ref{tab:semantic_segmentation}. Our approach can generate high-performing neural network parameters for the semantic segmentation task.

\begin{table}[htbp]
    \centering
    \tablestyle{10pt}{1.2}
    \begin{tabular}{y{50}x{20}x{35}}
         model &  original & pdiff  \\
        \shline
         FCN & 60.5& \textbf{60.7} \\
    \end{tabular}
    \caption{P-diff on the semantic segmentation task. Results are reported in mIOU on a subset of the COCO dataset.}
    \label{tab:semantic_segmentation}
\end{table}

\paragraph{Image generation.} In this task, we use the DDPM~\cite{ho2020denoising} model pretrained on the CIFAR-10 dataset. Note that we use p-diff to only generate the last convolutional layer parameters of UNet in DDPM. For evaluation, we compute the FID score of models generated by p-diff and report the result in Table \ref{tab:image_generation}. We observe that p-diff achieves competitive performance compared to the original DDPM models, demonstrating its applicability to the image generation task.

\begin{table}[h]
    \centering
    \tablestyle{10pt}{1.2}
    \begin{tabular}{y{50}x{20}x{35}}
         model& original & p-diff  \\
        \shline
         DDPM UNet & 3.17 & \textbf{3.19} \\
    \end{tabular}
    \caption{P-diff on the image generation task. Results are reported in FID on the CIFAR-10 dataset.}
    \label{tab:image_generation}
\end{table}

\subsection{Bottleneck to Scaling Up}
Below, we examine how the GPU memory cost of p-diff changes when training with larger model checkpoints. Note that p-diff involves two training stages: autoencoder and diffusion model. For each type of model checkpoint, we increase the latent dimension of the autoencoder based on the size of model checkpoints and record the GPU memory cost with a fixed batch size of 50. As shown in Table~\ref{tab:bottleneck of scaling up}, the memory cost of both the autoencoder and the diffusion model increases proportionally with the size of the model checkpoints. Training on model checkpoints with more than 300K parameters usually exceeds the 40GB memory limit of the NVIDIA A100 GPU.

\begin{table}[h]
\centering
\tablestyle{6pt}{1.2}
\begin{tabular}{lccc}
model parameter & AE (MB) &latent dimension & diffusion (MB) \\
\shline
2048   & 658  &128 & 530 \\
30016  & 3756  &256& 1418 \\
100288  & 13352 &512& 3594 \\
300288 & 44882 &1024& 12942 \\
\end{tabular}
\caption{The bottleneck to scaling up is GPU memory.}
\label{tab:bottleneck of scaling up}
\end{table}

\section{Comparison with Related Studies}
We compare our method, p-diff, to each of the following related studies.

G.pt~\cite{peebles2023learning} uses the diffusion process as an optimizer (from starting parameter $\theta$ to future parameter $\theta^*$). It takes a noised version of the future parameter $\theta^*$ and generates a denoised parameter, condition on $\theta$ and the current and target loss / error. In contrast, we train p-diff to directly generate high-performing parameters from noise without any conditioning techniques. Also, models generated by G.pt often fall short of the original models' performance. P-diff can match or surpass the accuracy of original models.

MetaDiff~\cite{zhang2023metadiff} uses the diffusion process to perform meta-learning. It takes the features and labels of all training data as inputs and learns to generate target model checkpoints from randomly initialized weights through a denoising process. However, p-diff is designed to generate high-performing neural network weights from random noise.

Hyper-representations~\cite{schurholt2022hyper} introduces a model zoo, containing numerous models trained from scratch. An autoencoder is then trained on the parameters of these models, enabling the generation of new models by sampling from the latent space and decoding the sampled latent vectors. Different from their approach, we use a diffusion model to model the latent distribution of high-performing models and generate novel models from random noise.

HyperTransformer~\cite{zhmoginov2022hypertransformer} is proposed to generate models for supervised and semi-supervised tasks where labeled data are necessary. It relies on image features to generate small target CNN parameters, while p-diff generates neural network parameters from noise. HyperTransformer is primarily compared with other few-shot learning methods, such as MAML~\cite{finn2017model}. In contrast, p-diff primarily compares the models it generates with those optimized using AdamW.

GHN methods~\cite{zhang2018graph, knyazev2021parameter, knyazev2023can} are originally designed for neural architecture search (NAS) training. These methods predict parameters with better weights through the given architectures as better initializations. Nevertheless, compared to p-diff, they are not able to synthesize model parameters that have comparable results with models trained using AdamW or other optimizers.

Denil \textit{et al.}~\cite{denil2013predicting} demonstrates that training only a small fraction (5\%) of a neural network's parameters can achieve performance comparable to, or even exceeding, that of the fully trained original model. This suggests that the remaining 95\% of the parameters are largely redundant and not critical for the network's training process.
In contrast, p-diff focuses on generating high-performing, novel models rather than the redundancy of parameters.

DWSNets~\cite{navon2023equivariant} introduces an approach to designing neural architectures that operate in deep weight spaces. They focus on what neural architectures can effectively learn and process neural models that are represented as sequences of weights and biases. However, p-diff emphasizes generating neural network parameters rather than designing neural architectures.

\end{document}